\documentclass[journal]{IEEEtran}
\pdfoutput=1

\usepackage[utf8]{inputenc}
\usepackage{cite}
\usepackage[pdftex]{graphicx}
\usepackage{xspace}
\usepackage{comment}
\usepackage{fixltx2e}
\usepackage[binary-units=true, detect-all=true]{siunitx}
\usepackage{multirow}
\usepackage{csquotes}
\usepackage{subfig}
\usepackage{pdfpages}

\graphicspath{{../graphics/}{res/}}

\newcommand{\labpet}{LabPET\leavevmode\hbox{$\rm {}^{TM}$}\xspace}
\DeclareSIUnit[number-unit-product = \text{~}]
  \coincidences{coincidences}

\begin{document}
\bstctlcite{BSTcontrol}

\title{Automatic Channel Fault Detection and Diagnosis System for a Small Animal APD-Based Digital PET Scanner}

\author{Jonathan~Charest,~\IEEEmembership{Student~Member,~IEEE,}
        Jean-François~Beaudoin,
        Jules~Cadorette,
        Roger~Lecomte,~\IEEEmembership{Senior~Member,~IEEE,}
        Charles-Antoine~Brunet,~\IEEEmembership{Member,~IEEE,}
        and~Réjean~Fontaine,~\IEEEmembership{Senior~Member,~IEEE}
\thanks{
This work was supported by grants from the Natural Science and Engineering Research Council of Canada (NSERC), the Fonds de Recherche du Québec – Nature et Technologies (FRQNT) and the Regroupement Stratégique en Microsystèmes du Québec (ReSMiQ).

J. Charest, C.-A. Brunet and R. Fontaine are with the Department of Electrical Engineering and Computer Engineering, Université de Sherbrooke, QC, Canada. Jonathan.Charest@USherbrooke.ca

J.-F. Beaudoin, J.~Cadorette and R. Lecomte are with the Sherbrooke Molecular Imaging Centre and the Department of Nuclear Medicine and Radiobiology, Université de Sherbrooke, QC, Canada.

© 2014 IEEE. Personal use of this material is permitted. Permission from IEEE must be obtained for all other uses, in any current or future media, including reprinting/republishing this material for advertising or promotional purposes, creating new collective works, for resale or redistribution to servers or lists, or reuse of any copyrighted component of this work in other works.
}}

\maketitle
\thispagestyle{empty}

\begin{abstract}
Fault detection and diagnosis is critical to many applications in order to ensure proper operation and performance over time. Positron emission tomography (PET) systems that require regular calibrations by qualified scanner operators are good candidates for such continuous improvements. Furthermore, for scanners employing one-to-one coupling of crystals to photodetectors to achieve enhanced spatial resolution and contrast, the calibration task is even more daunting because of the large number of independent channels involved. To cope with the additional complexity of the calibration and quality control procedures of these scanners, an intelligent system (IS) was designed to perform fault detection and diagnosis (FDD) of malfunctioning channels. The IS can be broken down into four hierarchical modules: parameter extraction, channel fault detection, fault prioritization and diagnosis. Of these modules, the first two have previously been reported and this paper focuses on fault prioritization and diagnosis. The purpose of the fault prioritization module is to help the operator to zero in on the faults that need immediate attention. The fault diagnosis module will then identify the causes of the malfunction and propose an explanation of the reasons that lead to the diagnosis. The FDD system was implemented on a \labpet avalanche photodiode (APD)-based digital PET scanner. Experiments demonstrated a FDD Sensitivity of \SI{99.3}{\percent} (with a 95\si{\percent} confidence interval (CI) of~[\num{98.7},~\num{99.9}])\unskip~for major faults. Globally, the Balanced Accuracy of the diagnosis for varying fault severities is \SI{92}{\percent}\unskip. This suggests the IS can greatly benefit the operators in their maintenance task.
\end{abstract}

\section{Introduction}
\IEEEPARstart{R}{ecent} studies have shown the importance of quality control testing (QC) to ensure proper performance of positron emission tomography (PET) scanners~\cite{Matheoud2011}. Intelligent fault detection and diagnosis (FDD) systems have demonstrated qualities that meet the requirements for large nuclear experiments~\cite{Atlas2008} and can significantly reduce the workload of QC personnel~\cite{Kazarov2012}. Therefore, to minimize the burden of frequent calibration and QC procedures on complex medical imaging devices such as PET scanners, an intelligent system (IS) for channel FDD was proposed~\cite{Charest}. Fault prioritization and diagnosis were missing from the previous work but it is crucial for complete QC testing. Additionally, the proposed IS could not learn from new faults preventing it from adapting to its environment and coping with new types of faults.

This paper proposes a fault prioritization and diagnosis system to support the aforedesigned IS for the \labpet scanner~\cite{Fontaine2009} with the goals of increasing fault detection efficiency, implementing fault prioritization and diagnosis to allow evaluation of the complete IS performance.

\section{Intelligent System}
\begin{figure}[ht]
  \begin{center}
    \includegraphics[width=0.3\textwidth, page=6]{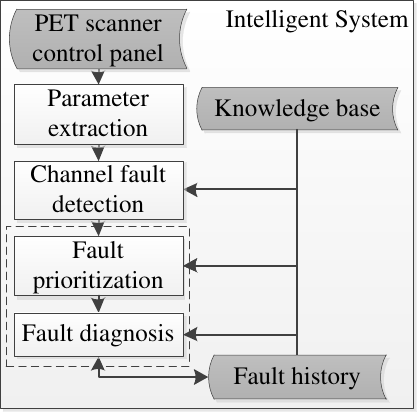}
  \end{center}
  \caption[Global architecture]{The global architecture of the intelligent FDD system features a modular architecture.}
  \label{fig:Architecture}
\end{figure}

The proposed IS system consists of 4 adaptable modules (\emph{parameter extraction}, \emph{fault detection}, \emph{fault prioritization} and \emph{fault diagnosis}) coupled to a knowledge base and a fault history database (Fig.~\ref{fig:Architecture}). The control panel of \labpet scanners provides data to the parameter extraction module, which transforms it to an appropriate format for the channel fault detection module. Then, the channel fault detection module generates a list of faulty channels that are afterwards sorted by the fault prioritization module. Finally, the fault diagnosis module produces a diagnosis for every channel in the fault list. The IS uses a knowledge base and a fault history database as prior information of the \labpet scanner.

\subsection{Fault Prioritization}
\begin{figure}[ht]
  \begin{center}
    \includegraphics[width=0.4\textwidth, page=7]{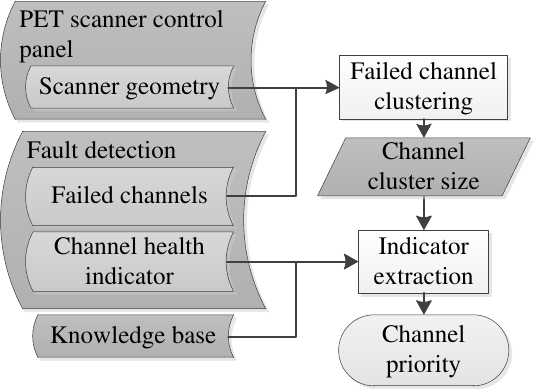}
  \end{center}
  \caption[Fault Prioritization]{The fault prioritization module uses fuzzy logic to extract the \emph{priority} indicator from the \emph{health} indicator and the proximity of other failed channels.}
  \label{fig:FaultPrioritization}
\end{figure}

\begin{figure}[ht]
  \begin{center}
    \subfloat[]{\label{fig:NemaReference}\includegraphics[width=0.24\textwidth]{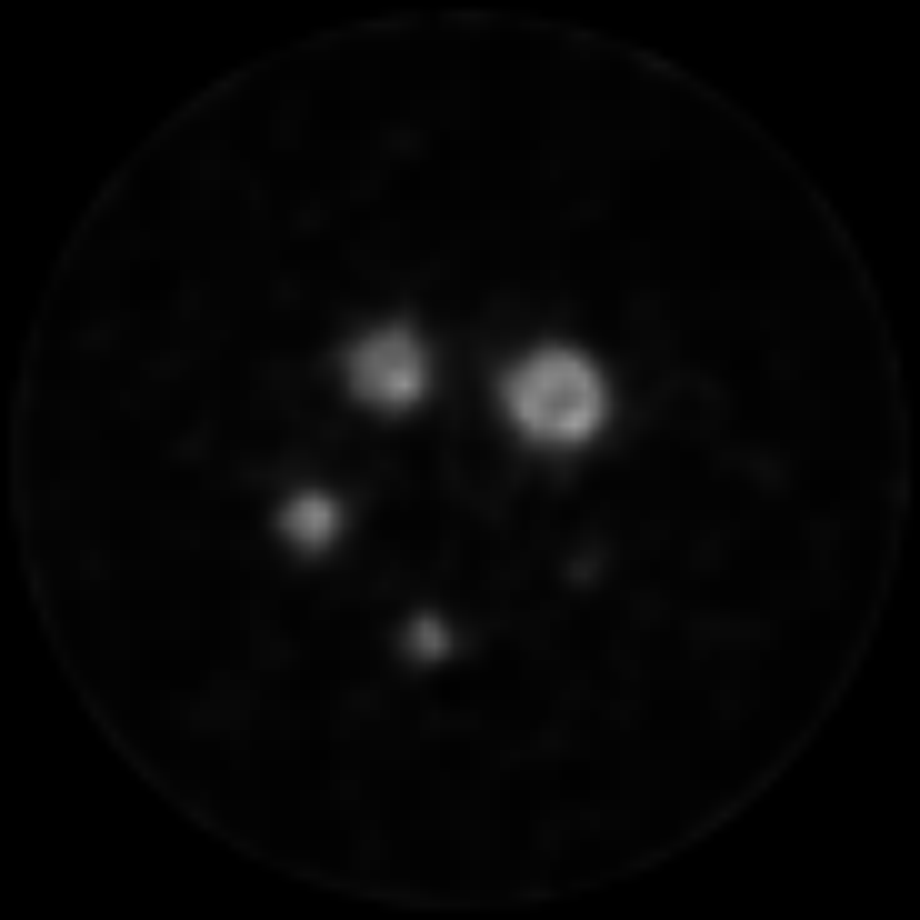}}
    \subfloat[]{\label{fig:NemaModified}\includegraphics[width=0.24\textwidth]{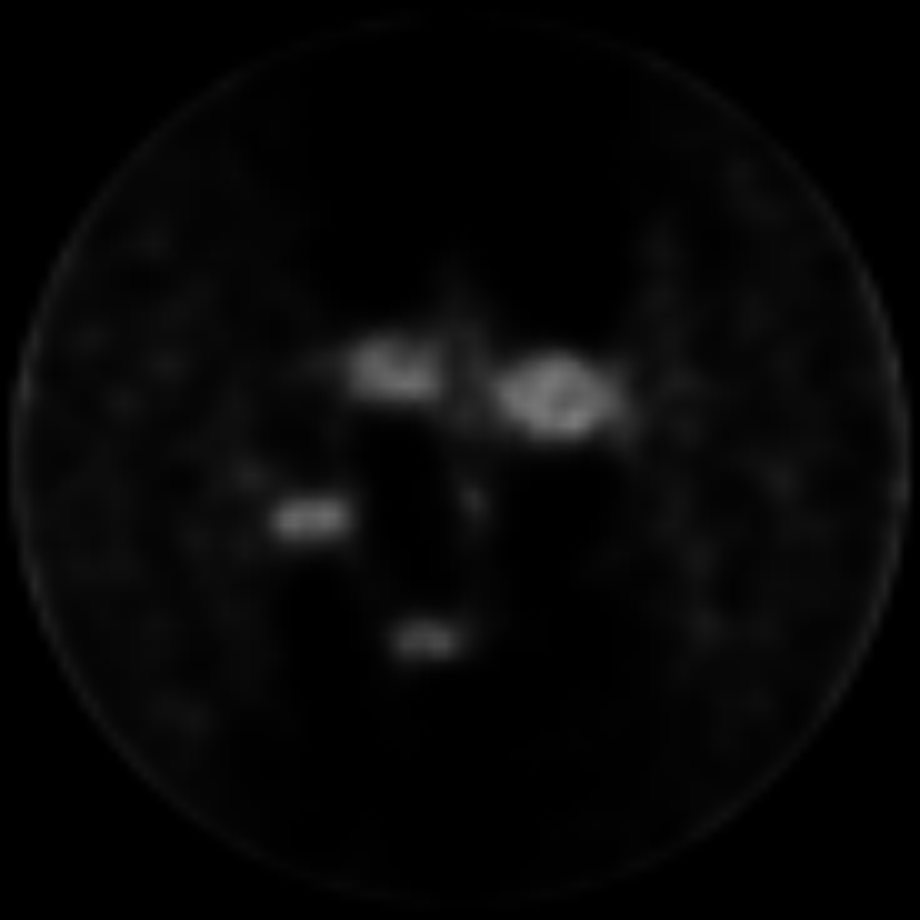}}
  \end{center}
\caption[Fault Prioritization]{Reconstruction of a NEMA image quality phantom with \num{256} channel failed out of \num{3072} (Fig.~\ref{fig:NemaModified}) compared to a reconstruction without these failures (Fig.~\ref{fig:NemaReference}).}
  \label{fig:NEMA}
\end{figure}

The purpose of the fault prioritization module is to help the operators to zero in on the faults needing immediate attention (Fig.~\ref{fig:FaultPrioritization}). To do this, the module ranks the detected channel faults by a \emph{priority} indicator determined from available data on individual channels. Since groups of nearby failed channels have a higher tendency of causing artifacts in reconstructed images, the \emph{priority} indicator consists of a channel \emph{health} indicator weighted by a clustering factor adjusted to the failed channel cluster size.
The fault prioritization module receives the list of channels to sort from the fault detection module. The \emph{priority} indicator is then extracted for all channels using fuzzy logic rules from the knowledge base. Clustering is performed by using the density-based spatial clustering of applications with noise (DBSCAN)~\cite{Ester1996} algorithm, which is adequate in this situation because of the number of fault clusters and shapes. The cluster size is mapped to a fuzzy logic variable with four linguistic terms (SMALL, MEDIUM, LARGE, HUGE) where a huge cluster consists of around \num{45} failed channels for the \labpet system. This value was determined by an experiment were multiple failed channel cluster sizes were simulated, and image quality parameters where extracted from the reconstructed images~\cite{NationalElectricalManufacturersAssociation2008}. A \num{45} channels cluster was found to reduce significantly the recovery coefficient of \SI{1}{\milli\meter} diameter sources, as well as almost doubling the uniformity percentage standard deviation in the reconstructed image of a uniform flood source. It thus makes it a good reference for a huge cluster term (one that needs immediate attention). The effects of a \num{45} channels cluster is hard to detect by visual inspection of the image, but Fig.~\ref{fig:NEMA} shows the effects on a NEMA image quality phantom of \num{256} channel failed out of \num{3072} in a \labpet scanner. Finally, the \emph{fuzzylite}~\cite{Rada-Vilela2013} fuzzy logic library was used to quickly implement this module in C++.

\subsection{Fault Diagnosis}
\begin{figure}[ht]
  \begin{center}
    \includegraphics[width=0.5\textwidth, page=8]{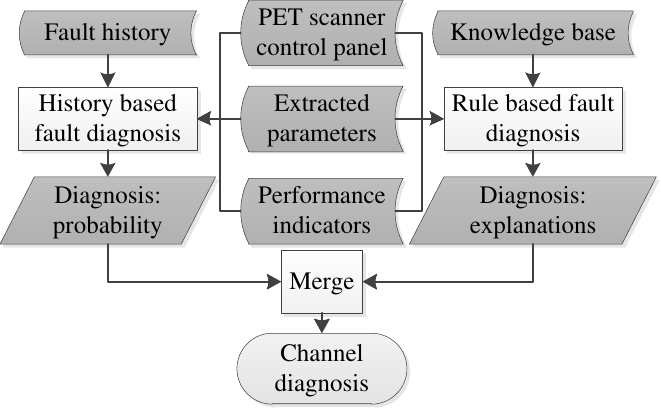}
  \end{center}
  \caption[Fault Diagnosis]{The fault diagnosis module uses a machine learning algorithm to identify the most probable diagnosis. A rule based diagnosis is used in parallel to provide explanations for the diagnosis.}
  \label{fig:FaultDiagnosis}
\end{figure}

The fault diagnosis module produces a diagnosis of the detected faults so that appropriate actions may be undertaken to correct the faults (Fig.~\ref{fig:FaultDiagnosis}). To help the operator interpret the diagnosis, it includes the probability of the selected diagnosis and also provides detailed explanations of the reasons leading to the diagnosis. The inputs to this module are \emph{PET scanner control panel data}, \emph{extracted parameters} and \emph{performance indicators} from the fault detection module. They are used to perform 2 types of diagnosis in parallel: a history based diagnosis and a rule based diagnosis. The history based diagnosis module makes use of machine learning techniques and has the capacity to adapt to new types of faults as well as to provide the probability of the diagnosis. It uses an ever expanding fault history database to learn how to perform diagnosis and adapt to the environment. On the other hand, the rule based expert system (ES) diagnosis module makes use of a knowledge base to diagnose and provide detailed explanations of the diagnosis. The results from both the history based diagnosis and the rule based diagnosis are then merged into a complete diagnosis that is presented to the scanner operator. For example, a diagnosis could be: \enquote{Increase Polarization (96\%): Channel has a calibration problem (channel LYSO photopeak drift is high), channel is weak (strength is low, identification is failed, energy is failed), channel is not saturated and polarization increase is safe.}. The merge is a logical OR that ensures the channel diagnosis incorporates the unique parameters of both diagnosis methods.

Machine learning is made by random decision forests~\cite{Ho1995}, which is an ensemble method that uses a multitude of decision trees. Decision trees have historically been used for FDD~\cite{Fenton2001} and random decision forests show improvements over using only one tree. Every node in a decision tree is a question on the input parameters and the branches represent the answers to those questions. The leaves at the bottom of the three indicate the identified diagnosis for the channel being diagnosed. In decision forests, a multitude of trees are generated from training data and the diagnosis of each of the trees are taken into account. In the learning phase, the trees are constructed from subsets of the training data and in the classification phase, the \enquote{votes} of each of these trees are used to determine the posterior probability of the different diagnosis options. The training data was acquired by introducing faults in the scanner configuration as will be further detailed in section~\ref{sec:MAM}. In addition to the initial training data, the fault history will be routinely expanded by the scanner operators by confirming or infirming the system's diagnosis in their maintenance runs.

Rule based ES's perform inference on the set of rules that form the knowledge base. The rules are evaluated by the ES inference engine and typically are of the form: \enquote{if A and B then C}. This method allows an expert to code his knowledge in a series of rules that can then be used to diagnose channels as well as to describe the diagnosis in a language common to scanner operators. Relevant uses of rule based ES for FDD are the TDAQ FDD system~\cite{Kazarov2012} of the ATLAS experiment~\cite{Atlas2008} and the HAL9000 system~\cite{DeCataldo2011} of the ALICE experiment~\cite{Alice2008}. For the current IS, the rules of the knowledge base were derived from the experience of \labpet QC operators by performing multiple meetings as well as by doing an in depth analysis of the scanner modules and their performance. The C Language Integrated Production System (CLIPS)~\cite{CLIPS} was used to integrate the ES in the diagnosis module.

\subsection{Improved Fault Detection}
In a previous paper~\cite{Charest}, the fault detection was performed by applying a threshold on an extracted indicator representing the channels' health (the same used in fault prioritization). As mentionned in the fault diagnosis section, the machine learning alorithm provides the probability of all diagnosis and uses, among many input parameters, the \emph{health} indicator. It was found that using a threshold on the diagnosis probability gave a better fault detection accuracy than using only the \emph{health} indicator and was thus adopted as the new method for fault detection. Hence, all channels are now prioritized and diagnosed prior to applying the threshold, and then, the thresholding function of the fault detection module is applied. Currently, the fault detection threshold is crossed when the probability for a diagnosis reaches \SI{70}{\percent}.

\section{Materials and Methods}
\label{sec:MAM}
Experimental measurements were conducted on a \SI{8}{\centi\meter} axial length \labpet scanner at the Sherbrooke Molecular Imaging Center~\cite{Bergeron2009}. The LabPET~8 scanner is an avalanche photodiode (APD)-based small animal PET imaging system having \num{3072} channels. The experiments were carried out in order to:
\begin{itemize}
\item{Evaluate the correlation of the \emph{priority} indicator with the severity of faults.}
\item{Evaluate the IS FDD Sensitivity for major faults and the diagnosis hypothesis test Balanced Accuracy for varying fault severities.}
\item{Evaluate the IS FDD Sensitivity for varying fault severities.}
\item{Evaluate the IS severity diagnosis Sensitivity.}
\end{itemize}

All the experiments consisted in introducing fake faults in the acquisition channels of the scanner by modifying APD bias voltage and noise threshold in the scanner configuration file. After the faults were introduced, the required data was taken from the scanner control panel, the IS was used to detect and diagnose the introduced faults and performance metrics were evaluated. Major channel faults were introduced on 800 randomly selected channels by lowering the APD bias voltage by \SI{50}{\volt}. Additionally, fault severities were introduced by generating 5 distinct levels of modifications on APD bias voltage and noise threshold (120 channels per level per fault type). APD bias fault levels range from \SIrange{5}{25}{\volt} by both postive and negative \SI{5}{\volt} increments. Finally, noise threshold fault levels range from \SIrange{5}{25}{ADC bins} by both postive and negative \SI{5}{ADC bins} increments.

The experiments were conducted to extract the following performance metrics of the system: channel priority indicator responsiveness, global diagnosis statistics, FDD Sensitivity per severity level, and finally, severity diagnosis Sensitivity.

\subsection{Channel Priority Indicator Responsiveness}
As the fault severity increases, the \emph{priority} indicator should also increase so that the highest severity faults are assigned the highest priority. To test this correlation, the distribution of the \emph{priority} indicator for each severity was extracted using the results from the introduced fault severities. The Spearman rank correlation between the \emph{priority} indicator and fault level was also evaluated.

\subsection{Global Diagnosis Statistics}
The IS global performance was evaluated using measurements of classification test statistics. The global FDD classification test Sensitivity was evaluated for the major faults. For the experiments, this is evaluated by dividing the number of correct diagnosis by the number of introduced faults:
\begin{equation}
  \text{Sensitivity} = \frac{\text{true positives}}{\text{positive conditions}}
  \label{eq:sensitivity}
\end{equation}

The global FDD Specificity was evaluated as the number of correctly identified working channels divided by the number of properly working channels:
\begin{equation}
  \text{Specificity} = \frac{\text{true negatives}}{\text{negative conditions}}
  \label{eq:specificity}
\end{equation}

The Balanced Accuracy is the mean of Sensitivity (Eq.~\ref{eq:sensitivity}) and Specificity (Eq.~\ref{eq:specificity})) and it was also evaluated because Sensitivity only accounts for the modified channels (positive conditions). Since, the plain Accuracy can conduct to inflating estimates due to the moderate class imbalance in the test data as there are significantly less faults than working channels, the Balanced Accuracy was used.

\begin{equation}
  \text{Balanced Accuracy} = \frac{\text{Sensitivity} + \text{Specificity}}{2}
  \label{eq:accuracy}
\end{equation}

\subsection{FDD Sensitivity Per Severity Level}
The FDD Sensitivity was evaluated for each fault severity and for APD bias voltage modifications as well as noise threshold modifications. As before, this is evaluated by dividing the number of correct diagnosis for a severity and fault type by the number of introduced faults of the specified severity and fault type. This should help determine what types of faults are harder to diagnose and should also result in an indication of the IS responsiveness.

\subsection{Severity Diagnosis Sensitivity}
In addition to providing the faults causes, the system can also be used to diagnose the faults severities. This is very useful for scanner operators since the fault severity can guide the magnitude of the  correction that is to be applied in order to correct a fault. The severity diagnosis Sensitivity (Eq.~\ref{eq:sensitivity}) can be evaluated by dividing the number of correctly diagnosed severities for a fault type by the number correctly diagnosed faults of the same type.

\section{Results}
\subsection{Channel Priority Indicator Responsiveness}
\begin{figure}[ht]
  \begin{center}
    \includegraphics[width=0.5\textwidth]{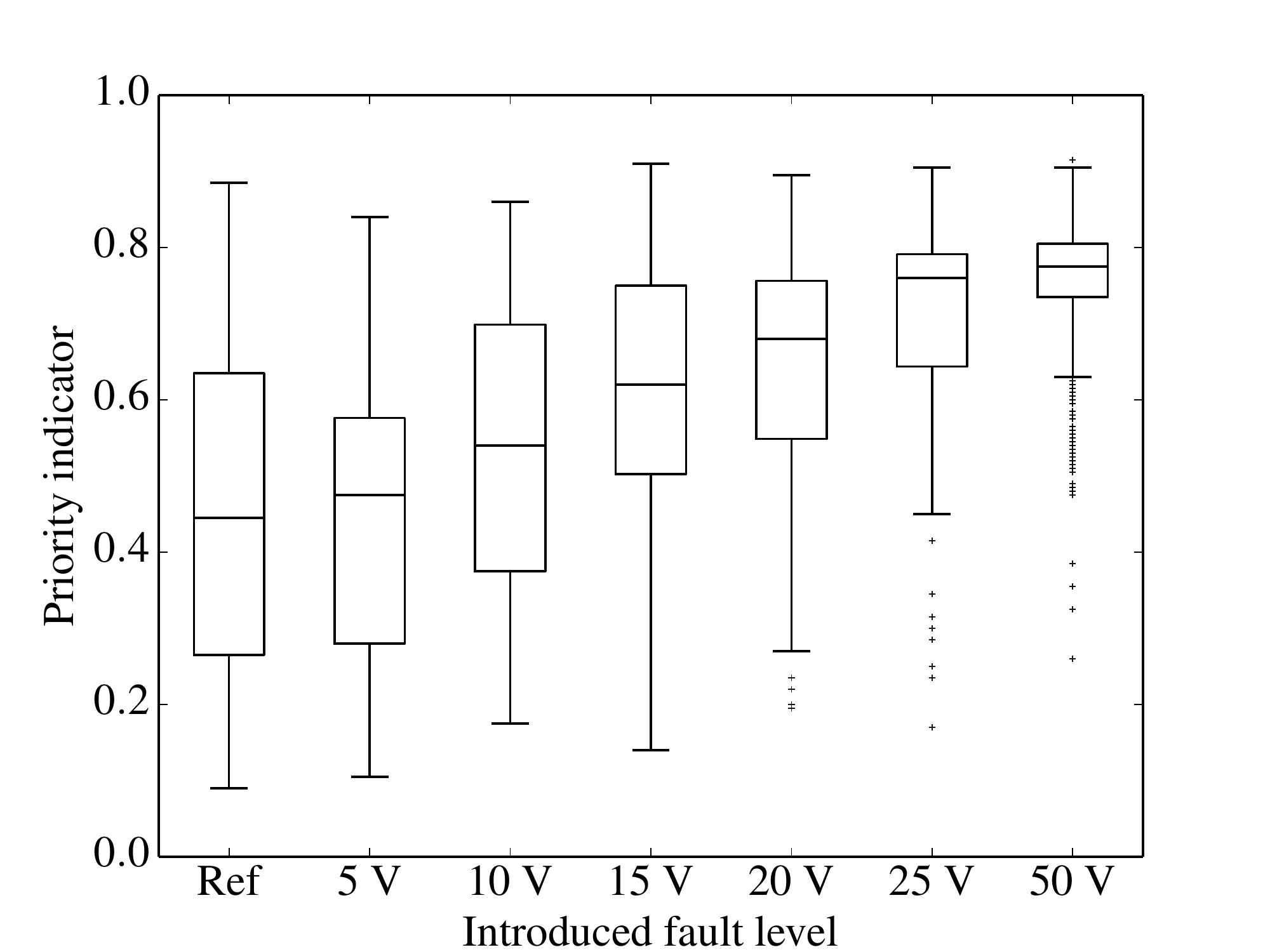}
  \end{center}
  \caption[Priority Polarization]{Boxplot of the channel \emph{priority} indicator for increasing APD bias fault severities. The boxes show the interquartile range (IQR) and outliers were identified at \SI{1.5}{IQR} from the median. The \enquote{Ref} label corresponds to the distribution of the \emph{priority} indicator before the introduction of faults.}
  \label{fig:PriorityPolarization}
\end{figure}

\begin{figure}[ht]
  \begin{center}
    \includegraphics[width=0.5\textwidth]{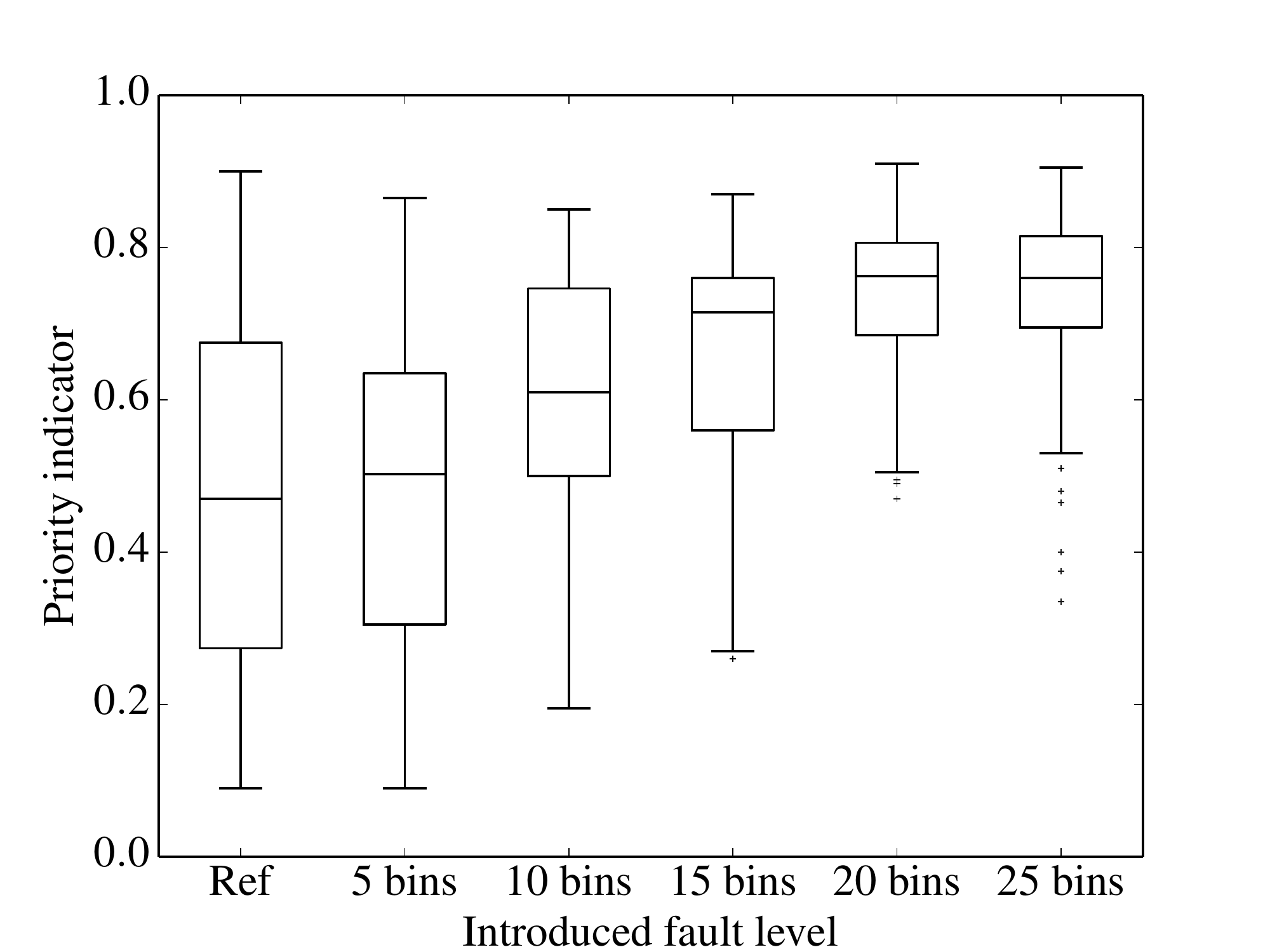}
  \end{center}
  \caption[Priority Noise Threshold]{Boxplot of the channel \emph{priority} indicator for increasing noise threshold fault severities.}
  \label{fig:PriorityNoiseThreshold}
\end{figure}

The distribution of the \emph{priority} indicator for each fault severity is shown in boxplots (Figs.~\ref{fig:PriorityPolarization} and \ref{fig:PriorityNoiseThreshold}). Boxplots better represent non Gaussian statistical distributions; they provide a good overview of non parametric distributions. In these plots, boxes show the interquartile range (IQR) and outliers where identified when data points are farther than \SI{1.5}{IQR} from the median. The IQR is evaluated from the quartiles ($\text{IQR} = \text{Q}_3 - \text{Q}_1$) so it is a measure of the distribution spread like the full width at half maximum (FWHM). Actually, for a Gaussian distribution both parameters are derived from $\sigma$ ($\text{IQR} \approx 1.349\sigma$ and $\text{FWHM} \approx 2.355\sigma$). The \enquote{Ref} column corresponds to the indicator distribution prior to the introduction of faults. The figures show a significant increase in priority when the severity increases. The Spearman rank correlation coefficient between the \emph{priority} indicator and the introduced fault severity is $\rho$=\num{0.36}, \textit{p}=\num{1.3e-19} for APD bias faults and $\rho$=\num{0.34}, \textit{p}=\num{1.1e-17}\unskip~for noise threshold faults, which indicates a fair correlation.

\subsection{Global Diagnosis Statistics}
The measured FDD Sensitivity for major faults (\SI{50}{\volt} APD bias decrease), which corresponds to the classification test Sensitivity, is \SI{99.3}{\percent} (CI:~[\num{98.7},~\num{99.9}])\unskip. The global diagnosis test Balanced Accuracy results for all introduced faults are shown in Table~\ref{tab:DiagnosisAccuracy}. It is important to keep in mind that a system randomly choosing a diagnosis would result in an Accuracy of \SI{50}{\percent}. At first glance, the IS global Balanced Accuracy remains fairly high irrespective of the fault type.
\begin{table}[ht]
  \renewcommand{\arraystretch}{1.3}
  \caption[Diagnosis Balanced Accuracy]{Global Diagnosis Test Balanced Accuracy}
  \label{tab:DiagnosisAccuracy}
  \centering
  \begin{tabular}{| c | c | c | c | c |}
    \hline
    \multicolumn{3}{| c |}{FDD Sensitivity (major faults):} & \multicolumn{2}{| c |}{\SI{99.3}{\percent} (CI:~[\num{98.7},~\num{99.9}])} \\
    \hline\hline
    \multicolumn{2}{| c |}{HV Bias} & \multicolumn{2}{c |}{Noise threshold} & \multirow{2}{*}{Globally} \\
    \cline{1-4}
    Increase & Decrease & Increase & Decrease & \\
    \hline
    \SI{95}{\percent} & \SI{87}{\percent} &
    \SI{91}{\percent} & \SI{88}{\percent} &
    \SI{92}{\percent}\\
    \hline
  \end{tabular}
\end{table}

\subsection{FDD Sensitivity Per Severity Level}
\begin{figure}[ht]
  \begin{center}
    \includegraphics[width=0.5\textwidth]{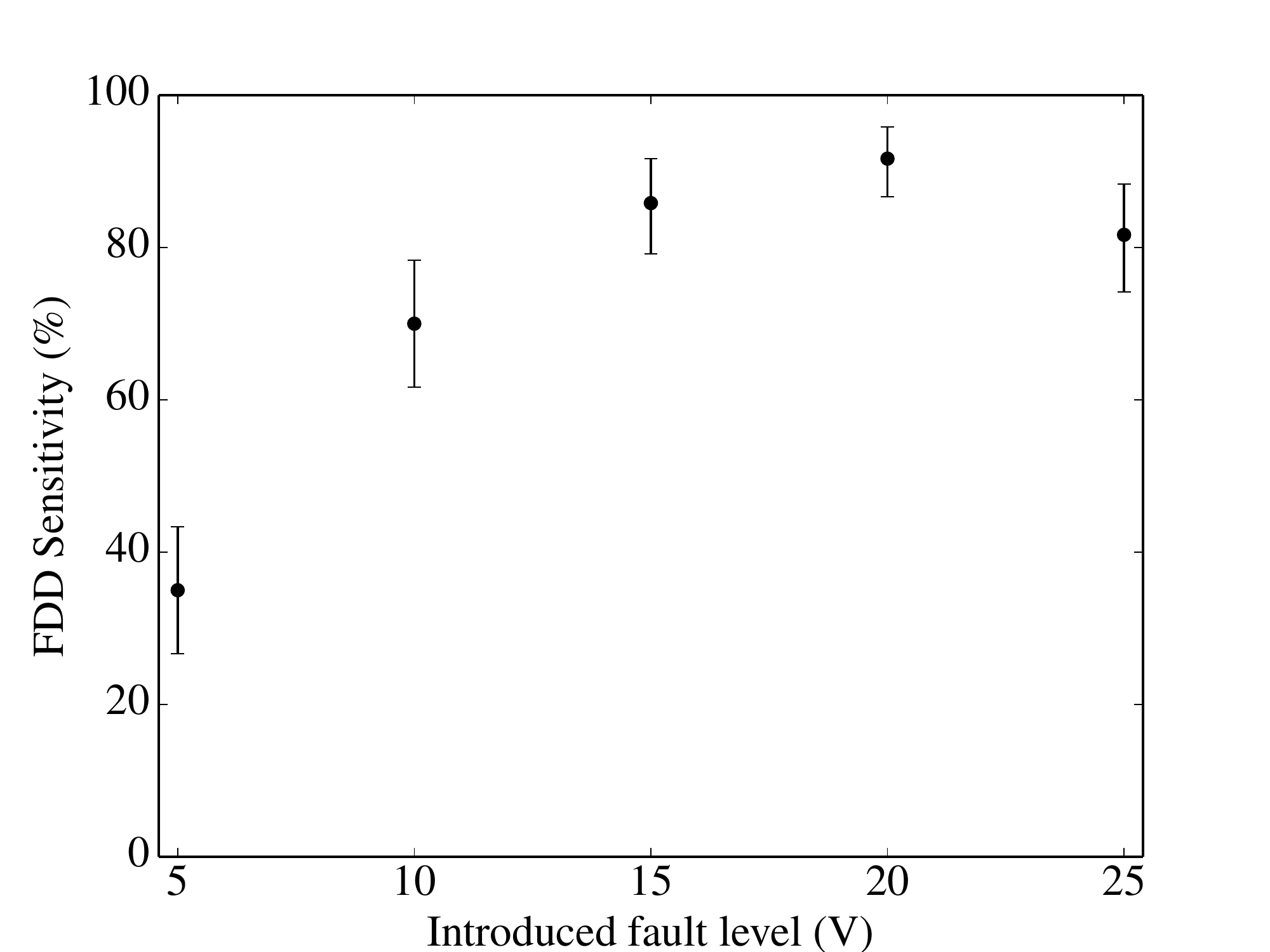}
  \end{center}
  \caption[FDD Sensitivity Polarization]{FDD Sensitivity for increasing APD bias fault severities.}
  \label{fig:FDDRatePolarization}
\end{figure}

\begin{figure}[ht]
  \begin{center}
    \includegraphics[width=0.5\textwidth]{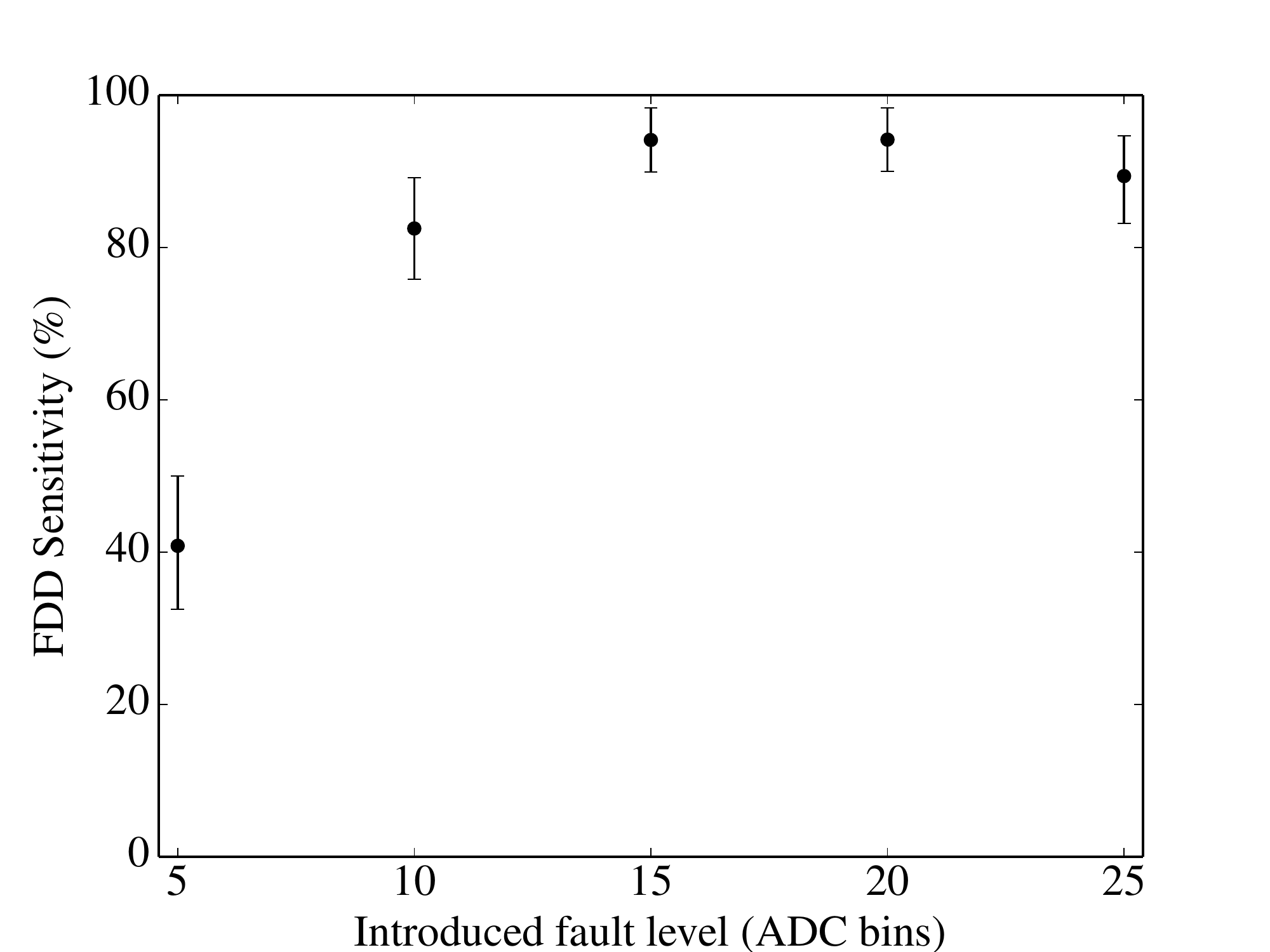}
  \end{center}
  \caption[FDD Sensitivity Noise Threshold]{FDD Sensitivity for increasing noise threshold fault severities.}
  \label{fig:FDDRateNoiseThreshold}
\end{figure}

The FDD sensitivities for varying fault severities are shown in Fig.~\ref{fig:FDDRatePolarization} for APD bias voltage modifications and Fig.~\ref{fig:FDDRateNoiseThreshold} for noise threshold modifications. As expected, FDD Sensitivity increases rapidly with fault severity. There is a slight drop in FDD Sensitivity for the last level for both fault types, an hypothesis for this observation will be provided in the discussion section.

\subsection{Severity Diagnosis Sensitivity}
\begin{figure}[ht]
  \begin{center}
    \includegraphics[width=0.5\textwidth]{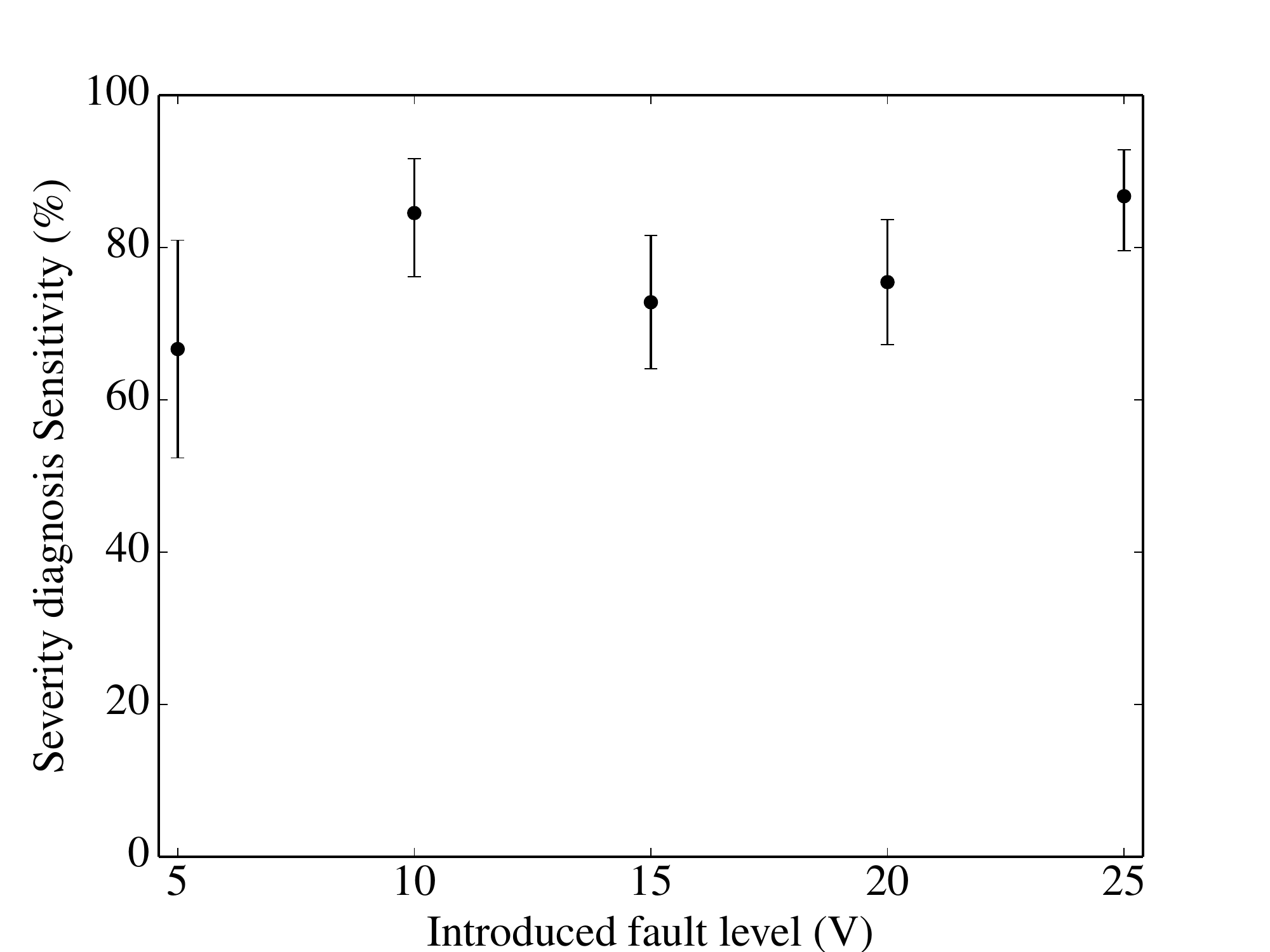}
  \end{center}
  \caption[Severity FDD Sensitivity Polarization]{Severity diagnosis Sensitivity for increasing APD bias fault severities.}
  \label{fig:FDDSeverityRatePolarization}
\end{figure}

\begin{figure}[ht]
  \begin{center}
    \includegraphics[width=0.5\textwidth]{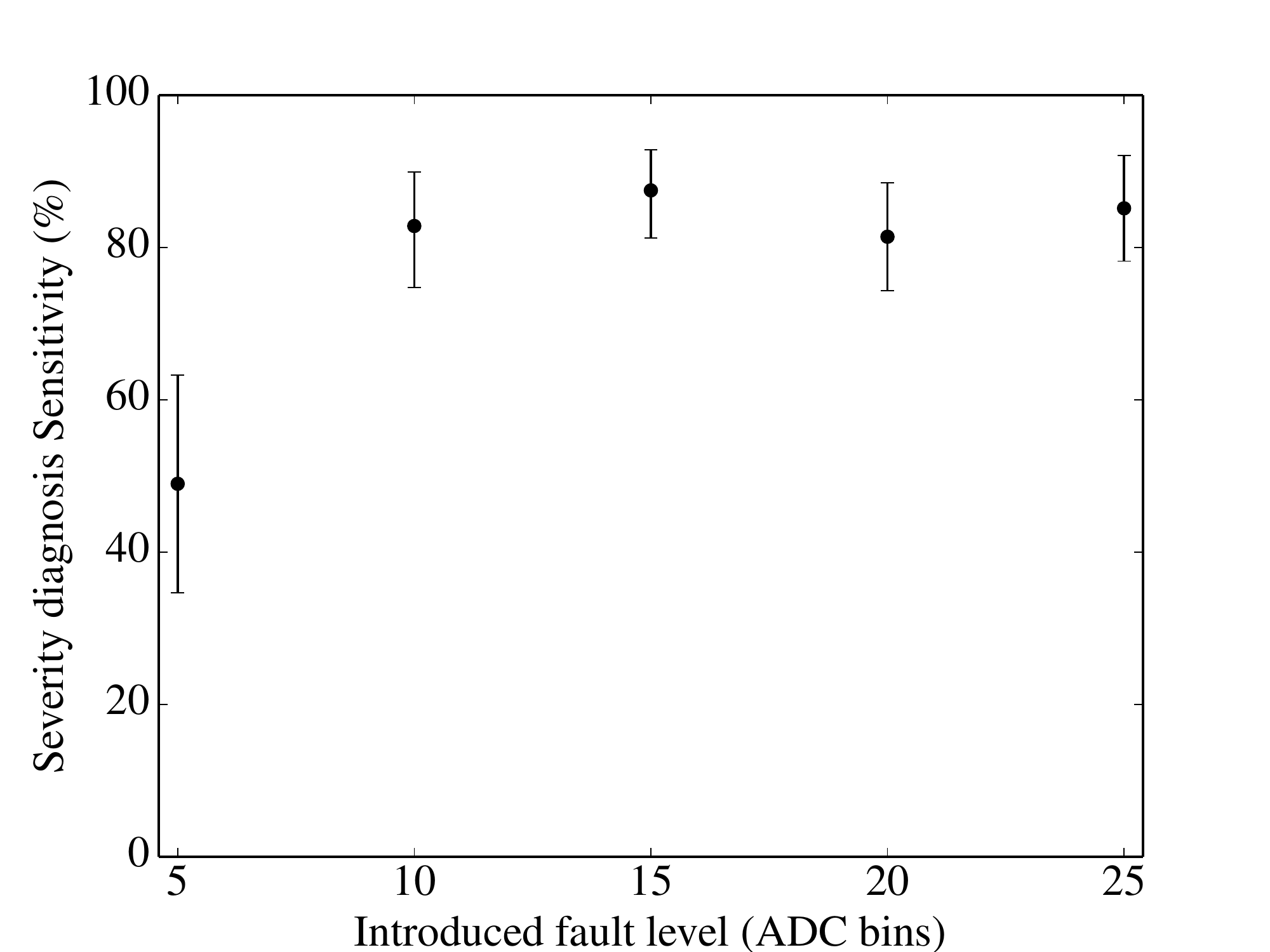}
  \end{center}
  \caption[Severity FDD Sensitivity Noise Threshold]{Severity diagnosis Sensitivity for increasing noise threshold fault severities.}
  \label{fig:FDDSeverityRateNoiseThreshold}
\end{figure}

Finally, the severity diagnosis is evaluated for different severities as shown in Figs.~\ref{fig:FDDSeverityRatePolarization} and \ref{fig:FDDSeverityRateNoiseThreshold}. The error bars are larger than for FDD Sensitivity since the severity diagnosis Sensitivity only takes into account the channels that had a correct diagnosis.


\section{Discussion}
There is a good correlation between the priority indicator and fault severity, which confirms that the indicator can be used to prioritize fault correction in the scanner. The global FDD Sensitivity of the system is excellent for major faults (\SI{50}{\volt}) and fair for different fault levels. Severity diagnosis Sensitivity is good and should be indicated to the scanner operators as it can help them determine the degree of the configuration modifications to apply. The Accuracy is acceptable even if some errors are to be expected, but it is worth mentioning that as the fault history expands, the results will likely improve. The results show that lower fault severities are not diagnosed reliably, but an important factor to consider is that lower levels of configuration change do not insure that faults are actually created and that they have any significant effect on image quality. Further investigation will be required to improve the robustness of the IS for mild to medium severity faults in the system.

An unexpected drop in FDD Sensitivity for the last level for both fault types was observed. This will need to be investigated further, but our hypothesis is that this level of fault triggers more changes in the channel \emph{performance indicators}, which are used by the fault diagnosis module to determine the fault diagnosis probability making multiple diagnosis plausible (higher probabilities). Currently, faults are detected when the probability for a diagnosis reaches \SI{70}{\percent}, so if two different diagnosis are plausible, even though one is higher than the other, this could lead to a missed detection. For major faults this situation seems to no longer apply as seen by the FDD Sensitivity rate.   

\section{Conclusion}
The fault prioritization and diagnosis modules of an intelligent system (IS) designed to perform channel fault detection and diagnosis (FDD) was proposed for the \labpet scanner. The modules were evaluated and have shown a FDD Sensitivity of \SI{99.3}{\percent} (CI:~[\num{98.7},~\num{99.9}])\unskip~for major faults and a Balanced Accuracy of \SI{92}{\percent}\unskip~for varying fault severities. The \emph{priority} indicator correlates well to channel fault severity, which indicates that the system will be able to sort faults effectively. The performance of the diagnosis module indicates that the IS is capable of diagnosing many faults automatically and is suitable for use in the field. Finally, the IS will continue to be investigated to improve performance and help reduce the burden of the quality control (QC) procedures for scanner operators.

\section*{Acknowledgment}
The author thanks the members of the Groupe de Recherche en Appareillage Médical de Sherbrooke (GRAMS) and the Sherbrooke Molecular Imaging Center (CIMS) for their cooperation in this project.

\newpage

\bibliographystyle{IEEEtran}
\bibliography{bst_control,sources_abrv,references}

\end{document}